\documentclass[runningheads]{llncs}

\usepackage{eccv}

\usepackage{eccvabbrv}

\usepackage{graphicx}
\usepackage{booktabs}

\usepackage[accsupp]{axessibility}  

\usepackage[breaklinks,colorlinks,citecolor=eccvblue]{hyperref}
\usepackage{graphicx}
\usepackage{amsmath}
\usepackage{amssymb}
\usepackage{booktabs}
\usepackage{bbm}

\usepackage{enumitem}

\usepackage{color}
\usepackage{orcidlink}
\usepackage{multirow}
\usepackage{tabularx}
\usepackage{bm}
\usepackage[normalem]{ulem}

\usepackage{algorithm2e}
\RestyleAlgo{ruled}

\usepackage{sidecap}
\usepackage{xcolor,colortbl}
\usepackage{float}

\usepackage{algorithm2e}
\RestyleAlgo{ruled}

\usepackage{sidecap}

\DeclareMathOperator*{\argmax}{arg\,max}

\useunder{\uline}{\ul}{}
\newcommand{\myparagraph}[1]{\vspace{3pt}\noindent{\bf #1}}

\begin{document}

\title{MTA-CLIP: Language-Guided Semantic Segmentation with Mask-Text Alignment}

\titlerunning{Language-Guided Semantic Segmentation with Mask-Text Alignment}

\author{Anurag Das \and
Xinting Hu \and
Li Jiang \and 
Bernt Schiele}

\authorrunning{A.~Das et al.}

\institute{Max Planck Institute for Informatics, Saarland Informatics Campus, Germany}
\maketitle

\begin{figure}
\centering
\includegraphics[width=0.9\linewidth]{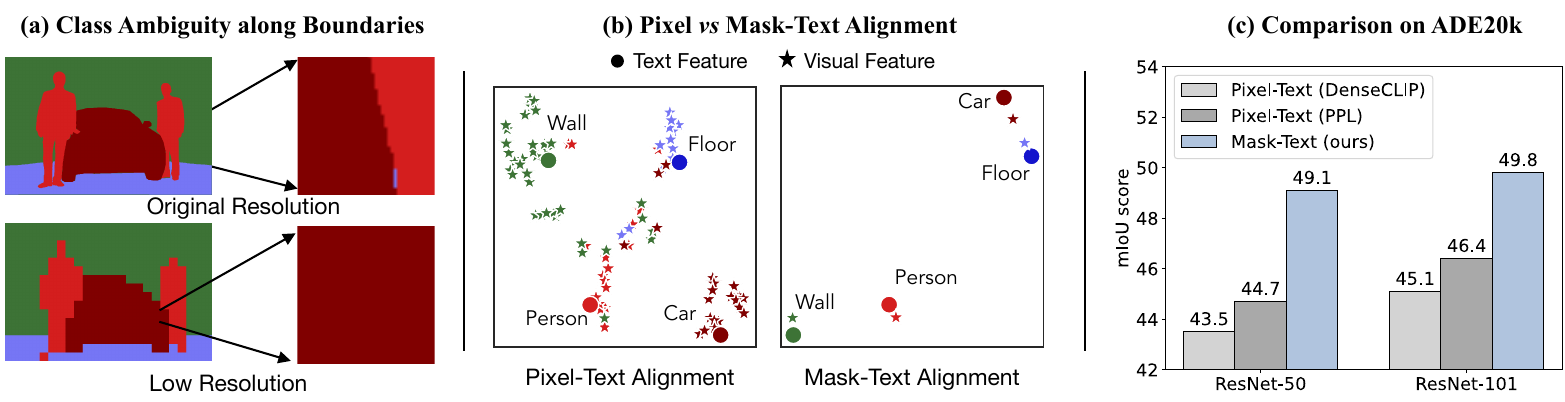}

\caption{(a): Pixel-Text Alignment relies on low-resolution image features for alignment and often encounters class boundary ambiguities. 
(b): t-sne visualisation for Pixel-Text \textit{vs.} Mask-Text Alignment. Pixel features of ``Wall'' are misaligned with the text feature of ``Person'' for Pixel-Text Alignment.  (c): Comparative results
on the ADE20k, showcase the superior performance of Mask-Text Alignment in semantic segmentation tasks.
on the ADE20k, showcase the superior performance of Mask-Text Alignment over Pixel-Text Alignment methods in semantic segmentation tasks.}
\label{fig:teaser1}
\end{figure}

\begin{abstract}    
Recent approaches have shown that large-scale vision-language models such as CLIP can improve semantic segmentation performance. 
These methods typically aim for pixel-level vision-language alignment, but often rely on low-resolution image features from CLIP, resulting in class ambiguities along boundaries.
Moreover, the global scene representations in CLIP text embeddings do not directly correlate with the local and detailed pixel-level features, making meaningful alignment more difficult.
To address these limitations, we introduce MTA-CLIP, a novel framework employing mask-level vision-language alignment. 
Specifically, we first propose Mask-Text Decoder that enhances the mask representations using rich textual data with the CLIP language model. Subsequently, it aligns mask representations with text embeddings using Mask-to-Text Contrastive Learning.
Furthermore, we introduce Mask-Text Prompt Learning, utilizing multiple context-specific prompts for text embeddings to capture diverse class representations across masks. 
Overall, MTA-CLIP achieves state-of-the-art, surpassing prior works by an average of 2.8\% and 1.3\% on standard benchmark datasets, ADE20k and Cityscapes, respectively.
\end{abstract}
 
\begin{figure}
\centering
\includegraphics[width=\linewidth]{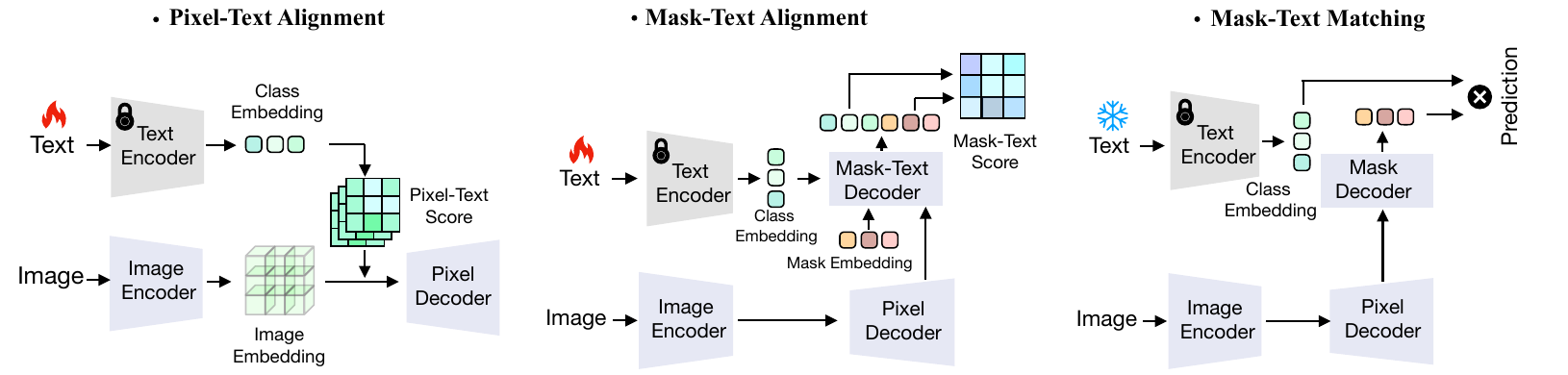}
\caption{Comparison between Pixel-Text Alignment, Mask-Text Alignment (ours) and Mask-Text Matching segmentation architectures. Flames and snowflakes refer to learnable and frozen parameters, respectively.
}
\label{fig:teaser2}
\end{figure}

\section{Introduction}
\label{sec:intro}

Semantic segmentation, a fundamental task in computer vision, has been significantly advanced by the integration of language models,  particularly with the advent of large-scale vision-language models (VLMs) such as CLIP~\cite{radford2021learning}. Recent studies~\cite{rao2022denseclip, kwon2023probabilistic}  
have demonstrated that pre-training with CLIP can considerably enhance the performance of semantic segmentation. 
The core strength of CLIP is its training on diverse datasets rich in visual and textual content, which enables a context-sensitive understanding of visual content. 
In semantic segmentation, the adoption of CLIP in training is defining new benchmarks, highlighting the benefits of cross-modal embeddings.

The common practice of integrating CLIP into semantic segmentation adopts pixel-text alignment~\cite{rao2022denseclip,kwon2023probabilistic}. In this approach, each pixel feature is aligned with the text embedding of its corresponding ground truth class label as illustrated in~\cref{fig:teaser2} (a).  
However, this approach faces several challenges. 
Firstly, due to computational constraints, pixel-text alignment often operates on low-resolution image features. This down-sampling of image features, caused by the network's image encoder stride and pooling, often results in the blurring of boundaries between different classes. This results in inaccurate alignment with the respective text embeddings.
As illustrated in~\cref{fig:teaser1} (a), a region containing a ``car'' and a ``person'' is indiscriminately labeled as ``car'' after down-sampling. As a result, the text embedding for ``car'' does not accurately align with the pixel-level image feature over that region. Moreover, CLIP performs image-text matching~\cite{gao2023clip}, resulting in text embeddings with global image-level semantics. It introduces mismatches when the text embeddings are aligned with the detailed and localized pixel features. As illustrated in~\cref{fig:teaser1} (b), certain pixel features for ``wall'' can be easily confused with the text embedding of ``person''. In contrast, when using mask features\footnote{we observe only one mask feature captures the entire class, for each of the classes in this sample. We provide more visualisations in the supplement.}
 for classes ``wall'' and ``person'', the mask features align more effectively with the corresponding text embeddings.

To address the challenges posed by pixel-text alignment, our MTA-CLIP introduces the Mask-Text alignment.
Unlike pixel representations, each mask representation with inherent boundary information captures entire entities or regions within a class. Such embeddings are more in line with CLIP's pre-training approach and avoid class ambiguity compared to per-pixel alignment.
The key component of MTA-CLIP is the novel Mask-Text Decoder. 
Different from Mask-Text matching (\cref{fig:teaser2}~\cite{ghiasi2022scaling,ding2021decoupling,xu2022simple,zhou2022extract,xu2023side}) that keeps the text embeddings fixed, our Mask-Text Decoder updates both mask and text embeddings. This enhances masks with rich textual data while enriching text embeddings with visual context using layer-wise attention mechanism. 
Furthermore, MTA-CLIP uses a contrastive learning approach within the decoder. This approach aligns the enhanced mask embeddings with corresponding text embeddings back into the CLIP feature space, leveraging the potential of CLIP's pre-trained model to regularize the mask representations.
Additionally, MTA-CLIP introduces Mask-Text Prompt Learning to effectively manage the diverse contexts within mask segments of the same class. This method utilizes distinct prompts for varying scenarios, thereby enhancing the model's adaptability and contextual awareness.

Overall, our MTA-CLIP achieves a Mask-Text alignment that  addresses the inherent limitations of pixel-text alignment. 
Our contributions are summarized as follows: 
\begin{itemize}[itemindent = 0pt,
              labelindent = \parindent,
              labelwidth = 2em,
              labelsep = 5pt,
              leftmargin = *]
    \item  We present MTA-CLIP, a novel CLIP based semantic segmentation framework that shifts from pixel-text to mask-text vision-language alignment, aligning more closely with CLIP’s holistic scene understanding and effectively resolving class ambiguities at boundaries.
    \item MTA-CLIP introduces Mask-Text Alignment comprising of two key components, Mask-Text Decoder, which enhances and aligns mask representations with rich textual data, and Mask-Text Prompt Learning, which employs multiple context-specific prompts to capture diverse class contexts within masks. 
    \item Extensive evaluation on standard datasets like ADE20k and Cityscapes shows that MTA-CLIP outperforms existing state-of-the-art methods, establishing new benchmarks in the semantic segmentation task with Mask-Text vision-language alignment.
\end{itemize}

\section{Related Work}

\myparagraph{Semantic Segmentation.} 
Semantic segmentation involves the task of assigning a categorical label to each pixel within an image, a challenge that has been extensively addressed using Deep Neural Networks (DNNs) ~\cite{krizhevsky2012imagenet, simonyan2014very, li2019selective, he2016deep}. Previous research has predominantly concentrated on two architectural paradigms, namely Convolutional Neural Networks (CNNs) and Transformer-based models. CNNs ~\cite{long2015fully,7913730, peng2017large, yu2020context, chen2017deeplab, zhao2017pyramid}, with their ability to capture local spatial hierarchies have shown to be highly effective. Noteworthy methods within this domain emphasize the utilization of multi-scale features, exemplified by DeepLab's Atrous Spatial Pyramid Pooling (ASPP)~\cite{chen2017deeplab,he2016deep}, PSPNet's Pyramid Pooling Module (PPM)~\cite{zhao2017pyramid}, Semantic Feature Pyramid Network (FPN)~\cite{kirillov2019panoptic} with its Feature Pyramid Network, incorporation of shape information as an additional prior~\cite{ding2019boundary,bertasius2016semantic,takikawa2019gated,yuan2020segfix}, and integration of high-resolution features~\cite{yu2021lite,sun2019high}. Conversely, transformers~\cite{xie2021segformer,strudel2021segmenter,wang2020axial,cheng2022masked}, distinguished by their capacity to capture global context, have exhibited superiority over CNN counterparts for semantic segmentation. While the majority of prior works have focused on per-pixel segmentation, assigning a label to each individual pixel, our work aligns with mask-based segmentation networks~\cite{cheng2022masked, cheng2021per}, aiming to partition an image into distinct masks with associated labels. Differently, we further enhance the mask representations with rich textual data following Mask-Text Alignment.

\myparagraph{Open Vocabulary Semantic Segmentation.}
To segment potentially unlimited classes, open vocabulary segmentation methods ``align'' pixel ~\cite{li2022languagedriven,cho2024cat} or mask features~\cite{ghiasi2022scaling,ding2021decoupling,xu2022simple,zhou2022extract,xu2023side} with textual representations. We refer open-vocabulary based ``alignment'' as ``matching'', given that the text features are not altered in such methods.
Pixel-Text Matching method~\cite{cho2024cat}, introduced a cost-aggregation that improves CLIP finetuning generalisation for open vocabulary segmentation. Mask-Text Matching (\cref{fig:teaser2}) methods ~\cite{xu2022simple,ding2023maskclip,ding2021decoupling,ghiasi2022scaling}, propose a two-stage framework, where mask proposals are generated in the first stage, and are matched with CLIP text features in the next stage. 
Differently, we focus on closed-set semantic segmentation, where our ``Mask-Text Alignment'' enhances both text and visual features, essential for optimal performance (see supplement Tab. 5). 
Further, we also learn multiple contextual prompts for text embeddings to best align diverse class representations across masks, whereas the prior works~\cite{xu2022simple, ding2023maskclip,ding2021decoupling,ghiasi2022scaling} do not use learnable prompts and use the handcrafted prompts such as ``a photo of'', ``a painting of'' instead. 

\myparagraph{Language Guided Semantic Segmentation.}
Recent works~\cite{rao2022denseclip, kwon2023probabilistic} have demonstrated that leveraging large-scale pretrained Vision-Language Models (VLMs) such as CLIP~\cite{radford2021learning} surpasses ImageNet-based approaches, attributed to their enhanced transferability to downstream tasks.
DenseCLIP~\cite{rao2022denseclip} initially proposed an extension of CLIP for downstream semantic segmentation tasks employing a pixel-text alignment method. Subsequently, PPL~\cite{kwon2023probabilistic} expanded upon DenseCLIP by introducing additional probabilistic prompts designed to capture diverse attributes in the data. Notably, both DenseCLIP~\cite{rao2022denseclip} and PPL~\cite{kwon2023probabilistic} rely on pixel-text vision language alignment, particularly at lower resolution features. Diverging from their approach, our proposal introduces a Mask-Text vision language alignment, specifically addressing challenges associated with pixel-text alignment (see~\cref{sec:intro}).

\myparagraph{Prompt Learning.}
Adapting Vision-Language Models (VLMs) to downstream tasks involves the finetuning of the model, a process that can result in the degradation of learned representations and potentially impact model transferability. Moreover, with an increase in model size, finetuning becomes computationally expensive. The concept of prompt learning, as introduced in prior studies such as~\cite{houlsby2019parameter, lester2021power, li2021prefix, khattak2023maple}, proposes an alternative approach by acquiring additional prompt tokens that tailor the large-scale VLM to specific downstream tasks. CoOp~\cite{zhou2022learning} optimizes the Contrastive Language-Image Pretraining (CLIP) model for downstream tasks by acquiring a set of input prompts for the VLM's language model. Building upon this, CoCoOp~\cite{zhou2022conditional} further refines the approach by incorporating image conditioning. Additionally, while the majority of existing methodologies concentrate on learning a singular prompt for improved generalization, ProDA~\cite{lu2022prompt} advocates for the utilization of multiple probabilistic prompts designed to cater to diverse contexts. In the context of semantic segmentation, DenseCLIP~\cite{rao2022denseclip} adopts a CoCoOp~\cite{zhou2022conditional} style of image-conditioned prompt learning for pixel-level predictions. Similarly, PPL~\cite{kwon2023probabilistic} extends the ProDA framework~\cite{lu2022prompt} by learning a probabilistic distribution of diverse prompts tailored to different image contexts. Both DenseCLIP and PPL employ pixel-text alignment to facilitate prompt learning. In contrast, our approach involves Mask-Text Alignment for the acquisition of distinct prompts catering to diverse contexts.

\begin{figure*}[t]

\centering
\includegraphics[width=\linewidth]{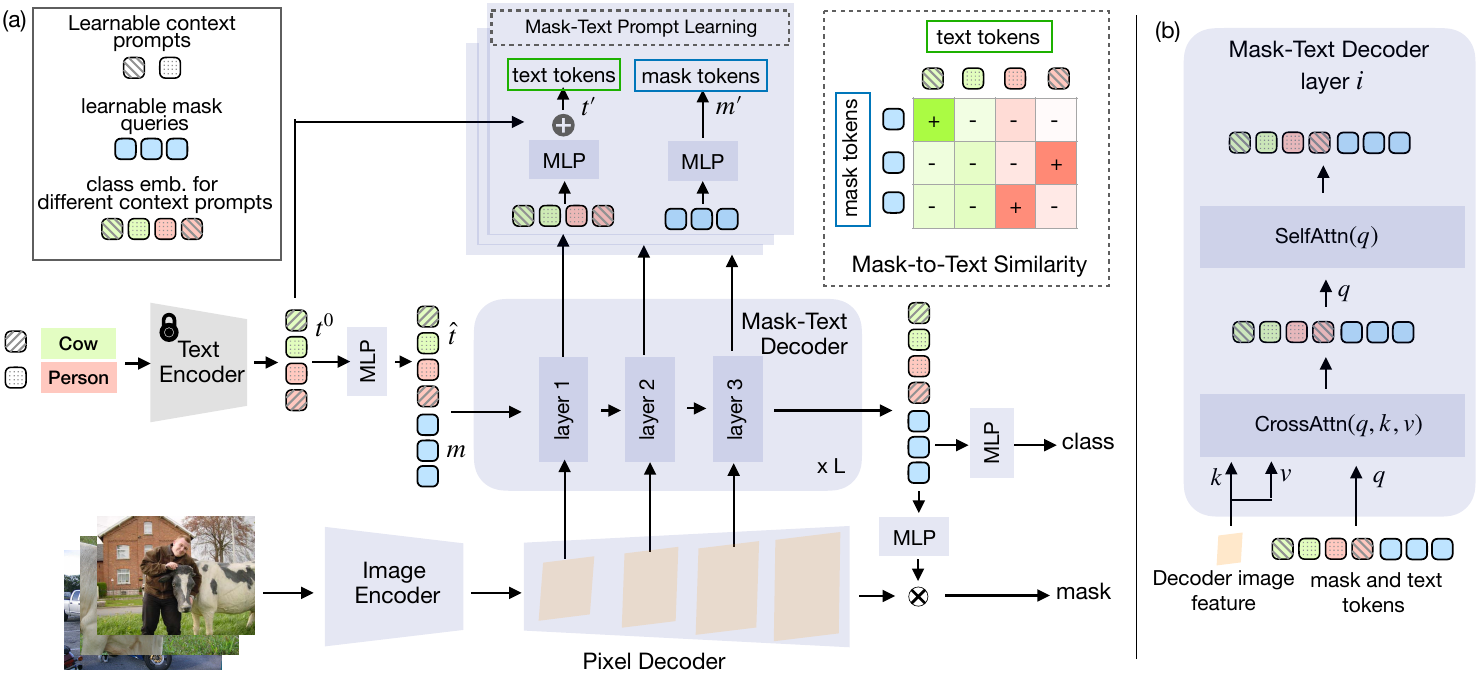}
\caption{Overview of our framework. (a) MTA-CLIP introduces Mask-Text Alignment (\cref{sub:mask-text}) that adopts a novel set of input queries combining both the mask queries ($\mathbf{m}$) and text embeddings ($\hat{\mathbf{t}}$). These combined queries interact with each other within mask-text decoder layers, enhancing mask representations through the incorporation of rich textual data following \textbf{Text-Enhanced Mask Feature Learning}. Subsequently, class-level alignment is conducted between the mask and text representations following \textbf{Mask-to-Text Contrastive Learning} via a contrastive loss applied to the mask-to-text similarity scores. (b) Furthermore, distinct alignment of different context prompts with mask representations is achieved using \textbf{Mask-Text Prompt Learning}.}
\label{fig:main}
\end{figure*}

\section{MTA-CLIP}
\label{sec:method}

In this section, we first describe the preliminaries (\cref{sec:prelims}), where we discuss mask-based segmentation method, Mask2Former~\cite{cheng2022masked} and Prompt Learning, as prior information required for our work. We then discuss the key component Mask-Text Alignment (\cref{sub:mask-text}), where we describe the two components, Mask-Text Decoder and Mask-Text Prompting, that effectively aligns the mask representations with the text embeddings. In the end, we present the training details (\cref{subsec:training}) for our framework.  

\subsection{Preliminaries}
\label{sec:prelims}
\myparagraph{Mask2Former.}
Mask2Former~\cite{cheng2022masked} is a mask-based segmentation method that segments an image into $N$ binary masks and classify them to $K$ classes. Different from standard pixel segmentation methods, it consist of an additional lightweight transformer decoder along with the segmentation backbone and a pixel decoder. The transformer decoder takes learnable tokens, $m_1, \dots, m_N$ (also called mask tokens) and pixel decoder features as input and processes these mask tokens with pixel decoder features with masked attention layers. It utilises pyramid resolution features extracted from the initial three layers of pixel decoder for processing mask tokens, whereas use the last layer (fourth layer) for making predictions. Typically, the dot product of the mask tokens with the last pixel decoder layer gives the final binary mask segmentations, whereas, an additional classification MLP predicts the class for these masks. Differently, in this work, our Mask-Text Decoder jointly process the text embedding and mask tokens. In particular, it enhances and align the mask tokens with CLIP text embeddings in different Mask-Text Decoder layers. Further, we also utilize multiple context prompts that captures diverse class representations among different masks. 

\myparagraph{Prompt Learning.} 
CLIP~\cite{radford2021learning} trained on 400M text-image pairs, learns powerful representations by aligning the image and text representations using contrastive learning. For better transferability on downstream tasks, prior works~\cite{rao2022denseclip,kwon2023probabilistic,zhou2022learning} learn contextual text prompts $\textbf{p}$, instead of using hand crafted prompts such as ``a photo of'', or ``a painting of''. The prompts $\textbf{p}$ are appended to the class token, $[\textbf{p}, \texttt{CLS}]$ and are fed to the frozen text encoder to obtain the context rich class embeddings $\mathbf{t} = [t_1, \dots, t_C] $ for $C$ classes. Different from ~\cite{rao2022denseclip,kwon2023probabilistic}, we learn prompts that align text features with mask representations. Moreover, we learn multiple context adaptive prompts to capture diverse class representations across different masks.

\subsection{Mask-Text Alignment}
\label{sub:mask-text}

The key component of MTA-CLIP is the novel Mask-Text Alignment mechanism, that combines class-specific language information from the CLIP text encoder with the mask representations. This alignment is achieved through two pivotal components: \textbf{(1) Mask-Text Decoder}, which effectively enriches and aligns mask representations with contextual language features to achieve more accurate and detailed segmentation, and  \textbf{(2) Mask-Text Prompt Learning}, which utilizes multiple context-specific prompts to capture the diverse representations of mask segments, thereby addressing the variability in real-world scenarios.

\subsubsection{1. Mask-Text Decoder.}
\label{subsub:mask-text-decoder}
Our Mask-Text Decoder has two key components: 1) Text-Enhanced Mask Feature Learning, where we use the rich textual data from language models to enhance mask representations, creating more context-aware segmentation; and 2) Mask-to-Text Contrastive Learning, aiming to establish a class-specific alignment between the mask representations and text embeddings within the CLIP feature space using a contrastive learning strategy. 

\myparagraph{Text-Enhanced Mask Feature Learning.}
\label{subsubsec:text-to-mask_rep}
In Text-Enhanced Mask Feature Learning, we enhance mask queries with class-specific language information derived from the CLIP text encoder. The process begins with constructing a novel set of queries $\mathbf{q}$, by concatenating mask queries $\mathbf{m}=[m_1, \dots, m_N]$ with text embeddings $\mathbf{t}=[t_1, \dots, t_C]$ for class labels from the CLIP text encoder, where $N$ denotes the number of queries and $C$ denotes the number of classes. Before concatenation, the text embeddings undergo dimension reduction through a MultiLayer Perceptron (MLP), adapting them to be compatible with the mask tokens in terms of dimension size and thus  enhances computational efficiency. This reduction is formulated as:
\begin{equation}
 \hat{\mathbf{t}} = \texttt{MLP}(\mathbf{t}), \quad \mathbf{t} \in \mathbb{R}^{C \times 1024}  \quad\text{and} \quad \hat{\mathbf{t}} \in \mathbb{R}^{C \times 256}. 
\end{equation}
Consequently, the resulting concatenated queries $\mathbf{q}$ are expressed as:
\begin{equation}
 \mathbf{q}=[\hat{\mathbf{t}}, \mathbf{m}], \quad \mathbf{q} \in \mathbb{R}^{(C+N) \times 256}. 
\end{equation}
Through this concatenation, $\mathbf{q}$ integrates both CLIP-based image features and text embeddings, serving as the input of the Mask-Text Decoder. 

Within the Mask-Text Decoder, which consists of $L$ layers, $\mathbf{q}$ undergoes a series of layer-wise attention refinements. At each layer $i \in \{0, 1,\cdots,L-1\}$, the process includes cross-attention and self-attention as follows:
\begin{align}
& \mathbf{q}_{\text{ cross}}^{(i)}= \texttt{CrossAttn}(\mathbf{q}^{(i)}, \mathbf{k}^{(i)}, \mathbf{v}^{(i)}),\\
& \mathbf{q}^{(i+1)} = \texttt{SelfAttn}(\mathbf{q}_{\text{ cross}}^{(i)})
\label{eq:attn}
\end{align}
Here, $ \mathbf{k}^{(i)}, \mathbf{v}^{(i)} $ are the $i^{\text{th}}$ level image features from the pixel decoder. During the cross-attention phase, both mask and text queries interact with image features, adapting to the specific context within the specific image. In the self-attention phase, mask queries not only interact among themselves to enhance the understanding of the relationships and context between different image areas, but also interact with text queries to integrate visual information with corresponding textual information. 

\myparagraph{Mask-to-Text Contrastive Learning.}
\label{subsubsec:mask-to-text-contrastive}
Building upon the Text-Enhanced Mask Feature Learning phase, where mask queries are augmented with visual and textual information, Mask-to-Text Contrastive Learning aims to establish class-level alignment between the mask and text representations within the CLIP feature space. This alignment is achieved through a contrastive learning approach, which involves enhancing the mask-text similarity score of a ``positive'' pair against those of ``negative'' pairs.

We first project $\mathbf{q}$, including both mask queries $\mathbf{m}$ and the text queries $\hat{\mathbf{t}}$, back to the CLIP feature space after each Transformer decoder layer. This projection is conducted using MultiLayer Perceptrons (MLPs):
\begin{equation}
\mathbf{t}' = \texttt{MLP}(\hat{\mathbf{t}})  + \mathbf{t}^0, \quad \mathbf{m}' = \texttt{MLP}(\mathbf{m})
\end{equation}
Here, the addition of the initial CLIP text embedding $\mathbf{t}^0$ to the projected text embeddings ensures the preservation of the original textual space from the CLIP model.

The essence of contrastive learning involves calculating the pairwise similarity between each projected mask query $m_i', (i \in \{1,\cdots, N\})$ and text query $t_j', (j \in \{1, \cdots, C\})$. The similarity score $S$, a dot product normalized by a temperature parameter $\tau$, is given by:
\begin{equation}
S(m'_i, t'_j) = \frac{m'_i \cdot t'_j}{\tau}.
\end{equation}
We select the text query with ground truth class for each mask query as the ``positive'' class, while ``negative'' pairs include the mask query and all other non-corresponding text queries.
The alignment is refined through a contrastive loss function, $\mathcal{L}_{sim}$, which is designed to maximize the similarity for the positive pair and minimize it for negative pairs. The loss is defined as:
\begin{scriptsize}
\begin{equation}
\begin{split}
\mathcal{L}_{sim} = - \frac{1}{N} \sum_{i=1}^N \sum_{j=1}^C \  \Big[  \  y_{ij} \cdot 
\log \frac{\exp\big( S(m'_i, t'_j)\big)}{ \exp\big( S(m'_i, t'_j)\big) + \sum\limits_{\substack{r \in \{1,\cdots, C\} \\r \ne j }} \exp\big( S(m'_i, t'_r)\big)}\Big]
\end{split}
\end{equation} 
\end{scriptsize}

Here, $y_{ij}$ equals $1$ for the mask-text pair where the text query belongs to the ground truth class for the given mask query, and $0$ otherwise. We obtain the ground truth class for the given mask query using bipartite matching of the predicted binary mask with actual ground truth masks similar to~\cite{cheng2022masked}.
This contrastive learning process aligns each mask query with its corresponding class text query. Moreover, we apply $\mathcal{L}_{sim}$ to $\mathbf{m}'$ and $\mathbf{t}'$ after each Mask-Text Decoder layer utilizing hierarchical feature pyramids from pixel decoder, enabling the model to accurately capture and segment across a range of mask sizes. 

\subsubsection{2. Mask-Text Prompt Learning.}
\label{subsubsec:context_adap}
Our approach tailors text prompts to distinctively align with the diverse contexts of mask segments within the same class. By implementing $K$ unique prompts shared across $C$ classes, we create an extensive range of $K \times C$ text queries from the text encoder. This expanded set of text queries is specifically designed to cater to the varied contexts in which a class can appear. By aligning each mask segment with this extensive array of class-context combinations, our model achieves a fine-grained Mask-Text alignment, leading to more precise and context-aware masks. To accommodate this variety in contextual representations, we expand the query set in Text-Enhanced Mask Feature Learning from $C+N$ to $K\times C+N$. For Mask-to-Text Contrastive Learning, we adopt a refined approach: For each mask token $m'_i$, the text query among the different $K$ instances from its ground truth class with the highest similarity score is identified as the ``positive'' query. Specifically, the positive text query for mask token $m'_i$ with ground truth class $j$ is defined as :
$  t'_{j,\widetilde{k}}, \text{ such that } \widetilde{k} = \argmax_{k\in \{1,\cdots,K\}} S(m'_i, t'_{j,k})$.
We employ two distinct strategies for selecting negative pairs:

\myparagraph{MixNeg}: All other text queries except the positive one (totaling $K\times C-1$) are treated as negative,  with the loss modified as:

\begin{scriptsize}
\begin{equation}
\label{eq:sim_context2}
\begin{split}
& \mathcal{L}_{sim}= - \frac{1}{N} \sum_{i=1}^N \sum_{j=1}^C \  \Big[  \  y_{ij} \cdot 
\log \frac{\exp \big( S(m'_i, t'_{j,\widetilde{k}})\big)}{ \exp \big( S(m'_i, t'_{j,\widetilde{k}})\big) + \sum\limits_{\substack{r \in \{1,\cdots C\}, k \in \{1,\cdots K\} \\ \{r,k\} \ne \{j,\widetilde{k}\}} } \exp\big( S(m'_i, t'_{r,k})\big)} \Big]
\end{split}
\end{equation} 
\end{scriptsize}

\myparagraph{SeperateNeg}: Text queries not belonging to the ground truth class (totaling $K\times (C-1)$) are treated as negative, with the loss modified as:
\begin{scriptsize}
\begin{equation}
\label{eq:sim_context}
\begin{split}
 &\mathcal{L}_{sim}=  - \frac{1}{N} \sum_{i=1}^N \sum_{j=1}^C \  \Big[  \  y_{ij} \cdot  \log \frac{\exp \big( S(m'_i, t'_{j,\widetilde{k}})\big)}{ \exp \big( S(m'_i, t'_{j,\widetilde{k}})\big) + \sum\limits_{\substack{r \in \{1,\cdots C\}, k \in \{1,\cdots K\}  \\  r \ne j}} \exp\big( S(m'_i, t'_{r,k})\big)} \Big]
\end{split}
\end{equation} 
\end{scriptsize}

As detailed in~\cref{pg:context_prompt}, both strategies exhibit improvement, with \textit{\textbf{MixNeg}} emerging as more effective due to its superior capability in differentiating between closely related contexts. 

\subsection{Training and Inference}
\label{subsec:training}
Our training process combines the strengths of both vision and language models to optimize the semantic segmentation performance of MTA-CLIP. The core of our training lies in the effective integration and alignment of mask representations with text embeddings through the Mask-Text Decoder. 

\myparagraph{Multi-Layer Mask-Text Alignment}: At each layer of the Mask-Text Decoder, we handle $N$ mask tokens alongside $K \times C$ class tokens derived from our Mask-Text Prompt Learning. The height and width of the pixel decoder image feature used at each Mask-Text Decoder layer, denoted as $h^{(i)}$ and $w^{(i)}$, vary depending on the layer. Similar to~\cite{cheng2022masked}, these dimensions are set as $(h^{(i)},w^{(i)})= (H/32, W/32),(H/16, W/16), (H/8, W/8)$ for layers $i~\texttt{mod}~3=0,1,2$, respectively, where $H$ and $W$ are the height and width of the original image, and $i \in [0,\cdots, 8]$ for Mask-Text Decoder with 9 layers. This variation in feature resolution across layers ensures Mask-Text Alignment at different scales, important for segmenting classes of different sizes.

\myparagraph{Overall Loss Function}: We adopt a composite loss function that balances mask classification, mask segmentation, and Mask-Text Alignment. Specifically, we use cross-entropy loss $\mathcal{L}_{mask-cls}$ for the classification prediction of mask tokens and binary cross-entropy loss $\mathcal{L}_{mask}$ for the segmentation prediction following~\cite{cheng2022masked}. Additionally, we have the similarity loss $\mathcal{L}_{sim}$ from Equation \ref{eq:sim_context2} for aligning mask tokens with context-specific class tokens. The final loss function is a weighted sum of these components:
\begin{equation}
\label{eq:sim_final}
\mathcal{L} = \lambda_1 \mathcal{L}_{mask-cls} + \lambda_2 \mathcal{L}_{mask} + \lambda_3 \mathcal{L}_{sim},
\end{equation}
where $\lambda_1$, $\lambda_2$, and $\lambda_3$ are weighting factors balancing the contribution of each loss component to the overall training objective.

\myparagraph{Inference}: During inference, we ignore the text tokens ($t_1,...,t_C$) and utilize only the mask tokens ($m_1,...,m_N$) from the final layer (9th layer) of the Mask-Text Decoder for final mask predictions. In particular, following~\cite{cheng2022masked} the text aligned mask tokens ($m_1,...,m_N$) are fed to an MLP layer, followed by dot product with pixel-decoder final layer features to give $N$ masks for the mask tokens. Moreover, these text aligned masks tokens goes through a separate classification MLP layer to obtain the class prediction for these mask tokens.

\section{Experiments}
In this section, we discuss the extensive experiments we conducted to demonstrate the effectiveness of MTA-CLIP. We compare our framework's performance with state-of-the-art methods for semantic segmentation. Then, we discuss ablation experiments followed by qualitative results.

\myparagraph{Datasets and Evaluation Metric.}
We evaluate MTA-CLIP on two standard dataset benchmarks i.e., ADE20k~\cite{zhou2019semantic} and Cityscapes~\cite{cordts2016cityscapes}. ADE20k has 150 categories, covering both indoor and outdoor scenes. It has around 20,000 images for training and 2,000 images for validation. Cityscapes, on the other hand, is an urban scene understanding dataset comprising 19 categories. It has 2,975 training images along with 500 validation images. Following prior works~\cite{rao2022denseclip,kwon2023probabilistic,cheng2022masked}, we report the mean Intersection over Union (mIoU) score for evaluation.

\myparagraph{Implementation Details.}
We employ Mask2Former~\cite{cheng2022masked} as our baseline segmentation network with CLIP pre-trained backbone for the image encoder. Following previous works~\cite{rao2022denseclip,kwon2023probabilistic}, we leverage all three available CLIP image backbones: ResNet-50~\cite{he2016deep}, ResNet-101~\cite{he2016deep}, and ViT-B~\cite{dosovitskiy2020image}. The text encoder is based on the CLIP-trained model, and we maintain its frozen state during training. In line with~\cite{rao2022denseclip}, we set the context length for each context prompt $\textbf{p}$ to 8. Moreover, we conduct experiments with $K=3$ different contextual prompts, unless otherwise specified.
Following prior works in the field~\cite{cheng2022masked,cheng2021per}, we set $N=100$ as the number of learnable mask tokens. Our final loss ($\mathcal{L}$) incorporates parameters $\lambda_1 = 2.0$, $\lambda_2 = 5.0$, and $\lambda_3 = 1.0$. Training hyperparameters, following~\cite{rao2022denseclip}, are as follows: for the ADE20k dataset, we utilize a crop size of $512\times512$, a batch size of 16, and train for 160k iterations. On the Cityscapes dataset, a crop size of $1024 \times\ 1024$, a batch size of 8, and 90k iterations are employed. The AdamW optimizer~\cite{loshchilov2017decoupled} is used with an initial learning rate of $2\times10^{-4}$.

{
 \setlength{\tabcolsep}{14pt}
 \renewcommand{\arraystretch}{1.0}
\begin{table}[t]
\caption{
Semantic Segmentation results on ADE20k dataset. We show comparisons of our method with prior works for different backbones. The mIoU (SS) and mIoU (MS) are single-scale and multi-scale inference results, respectively. 
$^*$: We re-implement Mask2Former with CLIP pre-trained backbones.}
 \centering
 \resizebox{\linewidth}{!}{%
\begin{tabular}{c|c|c|c|c}
\toprule
Backbone & Method & Pre-train & mIoU (SS) & mIoU (MS)  \\
\hline
\multirow{10}{*}{ResNet-50} & PSPNet~\cite{zhao2017pyramid} & ImageNet & 41.1  & 41.9  \\
\
 & DeepLabv3+~\cite{chen2018encoder} & ImageNet & 42.7 & 43.8  \\
 & UperNet~\cite{xiao2018unified} & ImageNet & 42.1 & 42.8   \\
 & Mask2Former~\cite{cheng2022masked} & ImageNet & 47.2 & 49.2  \\
  & Mask DINO~\cite{li2023mask} & ImageNet & 47.7 & -  \\
 & Semantic FPN~\cite{kirillov2019panoptic} & ImageNet & 38.6 & 40.6  \\
 & CLIP + Semantic FPN~\cite{radford2021learning} & CLIP & 39.6 & 41.6 \\
 & DenseCLIP~\cite{rao2022denseclip} & CLIP & 43.5 & 44.7   \\
 & PPL~\cite{kwon2023probabilistic} & CLIP & 44.7 & 45.8   \\
 & Mask2Former$^*$~\cite{cheng2022masked} & CLIP & 47.4 & 49.0  \\
 & \cellcolor[HTML]{EFEFEF}Ours & \cellcolor[HTML]{EFEFEF}CLIP &  \cellcolor[HTML]{EFEFEF}\textbf{49.1} &\cellcolor[HTML]{EFEFEF}\textbf{49.9}  \\

 \hline 
 \multirow{10}{*}{ResNet-101} & PSPNet~\cite{zhao2017pyramid} & ImageNet & 43.6 & 44.4 \\
\
 & DeepLabv3+~\cite{chen2018encoder} & ImageNet & 44.6 & 46.1   \\
 & UperNet~\cite{xiao2018unified} & ImageNet &  43.8 & 44.8  \\
 & Mask2Former~\cite{cheng2022masked} & ImageNet & 47.7 & 49.6  \\
& Semantic FPN~\cite{kirillov2019panoptic} & ImageNet & 40.4 & 42.3  \\
 & CLIP + Semantic FPN~\cite{radford2021learning} & CLIP & 42.7 & 44.3  \\
 & DenseCLIP~\cite{rao2022denseclip} & CLIP & 45.1 & 46.5   \\
 & PPL~\cite{kwon2023probabilistic} & CLIP & 46.4 & 47.8   \\
 & Mask2Former$^*$~\cite{cheng2022masked} & CLIP & 48.4 & 49.3  \\
 & \cellcolor[HTML]{EFEFEF}Ours & \cellcolor[HTML]{EFEFEF}CLIP &  \cellcolor[HTML]{EFEFEF}\textbf{49.8} & \cellcolor[HTML]{EFEFEF}\textbf{51.1}  \\
\hline
 \multirow{8}{*}{ViT-B} & SETR-MLA-DeiT~\cite{zheng2021rethinking} & ImageNet & 46.2 & 47.7   \\
 & Mask2Former~\cite{cheng2022masked} & ImageNet & 51.1 & 51.9  \\
    & BEiT~\cite{wang2023image} & ImageNet & 47.7 & -  \\
    & Semantic FPN~\cite{kirillov2019panoptic} & ImageNet-21K & 49.1 & 50.9  \\
 
  & CLIP + Semantic FPN~\cite{radford2021learning} & CLIP & 49.4 & 50.3  \\
 & DenseCLIP~\cite{rao2022denseclip} & CLIP & 50.6 & 51.3   \\
 & PPL~\cite{kwon2023probabilistic} & CLIP & 51.6 & 51.8   \\
 & Mask2Former$^*$~\cite{cheng2022masked} & CLIP & 51.0 & 51.9  \\
 & \cellcolor[HTML]{EFEFEF}Ours & \cellcolor[HTML]{EFEFEF}CLIP & \cellcolor[HTML]{EFEFEF}\textbf{52.3} & \cellcolor[HTML]{EFEFEF}\textbf{53.1} \\
\bottomrule 
\end{tabular}
}

\label{tab:main_result}
\end{table}
}

\subsection{Comparison with State-of-the-Art}

We compare MTA-CLIP's performance on two standard benchmarks i.e., ADE20k in~\cref{tab:main_result} and Cityscapes in~\cref{tab:city} with different backbones. We make two key comparisons. First, we compare our framework's performance with prior state-of-the-art's that uses VLM such as CLIP for semantic segmentation task. These methods~\cite{rao2022denseclip,kwon2023probabilistic} employ pixel-text alignment for aligning the vision and language modalities. Second, we compare the performance of our frameworks with an intuitive baseline of Mask2Former~\cite{cheng2022masked} with CLIP trained image encoder, without any Mask-Text Alignment. We also compare MTA-CLIP performance with classical semantic segmentation methods with ImageNet pre-training. Due to space constraint, we show Params and FLOPs comparison in supplement (Tab. 4). 

\myparagraph{Results for ADE20k.} MTA-CLIP exhibits superior performance over the state-of-the-art Pixel-to-Text alignment segmentation method PPL~\cite{kwon2023probabilistic}, achieving a performance gain of 4.4\% and 3.4\% for ResNet-50 and ResNet-101 backbones, respectively. This emphasizes the substantial impact of our Mask-Text Alignment approach compared to the Pixel-Text alignment approach for semantic segmentation tasks. Employing the ViT-B backbone, our framework achieves a performance of 52.3\%, surpassing PPL's 51.6\%. Furthermore, our method outperforms the prior state-of-the-art ImageNet pretrained Mask2Former by 1.9\%, 2.1\%, and 1.2\% for ResNet-50, ResNet-101, and ViT-B backbones, respectively. 

\myparagraph{Results for Cityscapes.} MTA-CLIP exhibits superior performance compared to the Pixel-to-Text alignment-based segmentation method, DenseCLIP~\cite{rao2022denseclip}, with gains of 2.2\% and 1.1\% for ResNet-50 and ResNet-101 backbones, respectively. Mask2Former with CLIP pretraining with our reimplementation improves over Mask2Former with ImageNet pretraining by 0.6\% and 1\% for ResNet-50 and ResNet-101 backbones, respectively. Our framework further surpasses this strong baseline by 0.7\% and 1.0\% for ResNet-50 and ResNet-101 backbones, respectively, showcasing the importance of Mask-Text alignment.

In summary, MTA-CLIP consistently outperforms previous methodologies in semantic segmentation by a substantial margin across standard dataset benchmarks. This observation shows the enhanced performance achieved through additional language alignment for masks. Additionally, our framework also exhibits superiority over previous semantic segmentation methods with ImageNet pre-training, emphasizing the significance of leveraging large-scale Vision-Language Models (VLMs) for network backbone pre-training. 

{
 \setlength{\tabcolsep}{2pt}
 \renewcommand{\arraystretch}{1.1}
\begin{table}[t]
 \caption{
Semantic Segmentation results on Cityscapes dataset.  We show comparisons of our method with two types of pretraining of image encoder (ImageNet and CLIP) for different backbones. $^*:$ We reimplement Mask2Former with CLIP pretrained backbone.}
 \centering
 \resizebox{\linewidth}{!}{%
\begin{tabular}{c|c|cc||c|c|cc}
\hline
\multicolumn{4}{c}{ResNet-50} & \multicolumn{4}{c}{ResNet-101} \\ 
\hline
 Method & Pre-train & mIoU (SS) & mIoU (MS) & Method & Pre-train & mIoU (SS) & mIoU (MS) \\ \hline
 PSPNet~\cite{zhao2017pyramid} & ImageNet & 78.5 & 79.8 &  PSPNet~\cite{zhao2017pyramid} & ImageNet & 79.8 & 81.0 \\
  DeepLabv3+~\cite{chen2018encoder} & ImageNet & 80.1 & 81.1   & DeepLabv3+~\cite{chen2018encoder} & ImageNet & 81.0 & 82.0 \\
  Mask2Former~\cite{cheng2022masked} & ImageNet & 79.4 & 82.2   & Mask2Former~\cite{cheng2022masked} & ImageNet & 80.1 & 81.9 \\
  UperNet~\cite{xiao2018unified} & ImageNet & 78.2 & 79.2 &   UperNet~\cite{xiao2018unified} & ImageNet & 79.4 & 80.4 \\
  DenseCLIP~\cite{rao2022denseclip} & CLIP & 79.2 & 80.3 &   DenseCLIP~\cite{rao2022denseclip} & CLIP & 81.0 & 81.8 \\
  Mask2Former$^*$~\cite{cheng2022masked} & CLIP & 81.0 & 82.1   & Mask2Former$^*$~\cite{cheng2022masked} & CLIP & 81.1 & 82.4 \\
  \cellcolor[HTML]{EFEFEF}Ours & \cellcolor[HTML]{EFEFEF}CLIP & \cellcolor[HTML]{EFEFEF}\textbf{81.7} & \cellcolor[HTML]{EFEFEF}\textbf{82.8}   & \cellcolor[HTML]{EFEFEF}Ours & \cellcolor[HTML]{EFEFEF}CLIP & \cellcolor[HTML]{EFEFEF}\textbf{82.1} & \cellcolor[HTML]{EFEFEF}\textbf{83.6} \\ \hline
\end{tabular}
}
\label{tab:city}
\end{table}
}

\subsection{Ablation Analysis}
\myparagraph{Ablation with different components.}
We conduct ablation studies on the components of our network, as detailed in~\cref{tab:ablation1}. `Baseline' denotes Mask2Former trained with the CLIP ResNet-50 backbone. We first enhance mask representations following Text-Enhanced Mask Feature Learning (Text-Enhanced,~\cref{subsubsec:text-to-mask_rep}) resulting in a performance gain of 0.6\% mIoU over baseline. Subsequently, by aligning mask representations with text representations with similarity loss using Mask-to-Text Contrastive Learning (MTCL,~\cref{subsubsec:mask-to-text-contrastive}), we observe a further increment of 0.5\% in mIoU. Finally, we use Mask-Text Prompt Learning (MTPL,~\cref{subsubsec:context_adap}) with $\textit{MixNeg}$ to learn distinct prompts, yielding an additional boost of 0.5\% in mIoU. The Mask-Text Decoder, which combines the learning of enhanced mask representations (Text-Enhanced) followed by aligning them with text representations (MTCL), gives a significant boost of 1.2\% in the mIoU score compared to the baseline, showing the importance of such alignment. 

\myparagraph{Results with other backbones.}
We extend our evaluation by comparing the performance of MTA-CLIP with alternative backbones, as presented in~\cref{tab:ablation1}. Specifically, we utilize ImageNet~\cite{deng2009imagenet} pretrained ViT-S~\cite{radford2021learning} and MiT-B4~\cite{xie2021segformer} backbones for this experiment. We demonstrate the versatility of our framework, indicating its applicability beyond Vision-Language Model (VLM)-trained backbones to also include ImageNet trained alternatives. In the case of ViT-S, DenseCLIP~\cite{rao2022denseclip} based on pixel-text alignment achieves of 47.8\% mIoU. Our MTA-CLIP framework, relying on Mask-Text Alignment, surpasses DenseCLIP by 3.6\% in mIoU and also outperforms the strong baseline Mask2Former by 0.8\%. Moreover, our framework exhibits effective compatibility with Mix Transformer backbones~\cite{xie2021segformer}, demonstrating a performance improvement of 1.4\% compared to SegFormer~\cite{xie2021segformer}, which employs a similar backbone. Collectively, these results affirm the adaptability of our framework across diverse backbones.

{
 \setlength{\tabcolsep}{6pt}
 \renewcommand{\arraystretch}{0.9}
\begin{table}

\caption{Left: Sementic Segmentation result with other backbones (ViT-S and MiT-B4). The experiments are performed on ADE20k dataset. Right: Ablation with different components of our method. Initially, we refine the mask representations using Text-Enhanced Mask Feature Learning (Text-Enhanced,~\cref{subsubsec:text-to-mask_rep}), followed by aligning the mask with text representations with Mask-to-Text Contrastive Learning (MTCL,~\cref{subsubsec:mask-to-text-contrastive}). Subsequently, we learn distinct context prompts using Mask-Text Prompt Learning (MTPL~\cref{subsubsec:context_adap}). The term `Baseline' refers to Mask2Former~\cite{cheng2022masked}. }
\centering
 \resizebox{\linewidth}{!}{%

\begin{tabular}{c|c|c}
\hline
Backbone & Method  & mIoU  \\
\hline
\multirow{4}{*}{ViT-S~\cite{dosovitskiy2020image}} &  Segmenter~\cite{strudel2021segmenter}  & 46.2   \\ 
 & DenseCLIP~\cite{rao2022denseclip}  & 47.8   \\
  & Mask2Former~\cite{cheng2022masked}  & 50.6   \\
  & \cellcolor[HTML]{EFEFEF}Ours & \cellcolor[HTML]{EFEFEF}\textbf{51.4} \\
\hline 
\multirow{3}{*}{MiT-B4\cite{xie2021segformer}} &  SegFormer~\cite{xie2021segformer}  & 50.3  \\ 
&  SegFormer (mask decoder)  & 51.2   \\ 
 & \cellcolor[HTML]{EFEFEF}Ours  & \cellcolor[HTML]{EFEFEF}\textbf{51.7} \\ 
\hline 
\end{tabular}

\quad
\begin{tabular}{l|c}
\hline
 Methods & mIoU \\
\hline 
Baseline & 47.4 \\
\hline 
+ Text-Enhanced & 48.1 \\
+ Text-Enhanced + MTCL & 48.6 \\
 \cellcolor[HTML]{EFEFEF}+ Text-Enhanced + MTCL + MTPL &  \cellcolor[HTML]{EFEFEF}\textbf{49.1} \\
\hline
 \end{tabular}
 }

\label{tab:ablation1}
\end{table}
}

\myparagraph{Enhancing Mask with Textual Data.} 
In this analysis (\cref{tab:ablation2}), we underscore the significance of enhancing mask representations with textual data from the pre-trained CLIP language model. Specifically, we modify the SelfAttn operation outlined in~\cref{eq:attn} to ensure that the text embeddings $\hat{\textbf{t}}$ and mask queries $\textbf{m}$ do not attend to each other. Instead, individual SelfAttn operations are applied to the text embeddings $\hat{\textbf{t}}$ and mask queries $\textbf{m}$, denoted as $\hat{\textbf{t}} = \text{SelfAttn}(\hat{\textbf{t}})$ and $\textbf{m} = \text{SelfAttn}(\textbf{m})$. It is important to note that, in this experiment, only the SelfAttn operation is modified, while all other operations remain the same. The results reveal a performance decline of 0.5\% when mask queries $\textbf{m}$ are not attended by the text embeddings $\hat{\textbf{t}}$. This observation shows the importance of attending mask queries with text embeddings for optimal performance.

\myparagraph{Comparison between Contextual Prompts.}
\label{pg:context_prompt}
As discussed in~\cref{subsubsec:context_adap}, we employ two distinct strategies, namely \textbf{\textit{SeparateNeg}} and \textbf{\textit{MixNeg}}, to derive specialized text prompts aligned with the diverse contexts of masks for a given class. In this experimental analysis, we first compare the efficacy of the two strategies, \textit{SeparateNeg} and \textit{MixNeg}. Subsequently, we assess our framework's performance across different numbers of text prompts ($K$). Overall, both approaches show effectiveness (see~\cref{tab:ablation2}), with \textit{MixNeg} exhibiting a performance gain of 0.5\%, compared to 0.3\% achieved by \textit{SeparateNeg}, over the baseline that utilizes a single prompt. Furthermore, our observations indicate that employing $K=3$, involving three learnable unique prompts shared across all classes, yields the optimal performance of 49.1\% with \textit{MixNeg}. Similarly, for \textit{SeparateNeg}, the best performance is achieved with $K=4$ across varying values of $K$.
{
 \setlength{\tabcolsep}{4pt}
 \renewcommand{\arraystretch}{1.0}
\begin{table}
\caption{Left: Importance of enhancing mask with rich text data. Experiments on ADE20k dataset. Right: Ablation with contextual prompts. In this experiment, we compare the two different approaches for effectively using multiple prompts as described in.... `\# prompts' refers to the number of different prompts used as input to CLIP text encoder.}
\centering
 \resizebox{\linewidth}{!}{%

\begin{tabular}{l|c}

Baseline & 47.4 \\
\hline 
+ Separate attentions in text and mask queries & 48.6 \\
+ \cellcolor[HTML]{EFEFEF} Joint attention in text and mask queries &  \cellcolor[HTML]{EFEFEF}49.1 \\
\hline
\end{tabular}

\quad
\begin{tabular}{l|c|c|c|c}
  \# prompts & 2 & 3 & 4 & 5 \\
 \hline 
 SeparateNeg & 48.5 & 48.7 & \textbf{48.9} & 48.2 \\
  \cellcolor[HTML]{EFEFEF}MixNeg  &  \cellcolor[HTML]{EFEFEF}48.6 &  \cellcolor[HTML]{EFEFEF}\textbf{49.1} &  \cellcolor[HTML]{EFEFEF}48.9 &  \cellcolor[HTML]{EFEFEF}48.4\\
 \hline
 
\end{tabular}
 }

\label{tab:ablation2}
\end{table}
}

\begin{figure}[t]
\centering
\includegraphics[width=\linewidth]{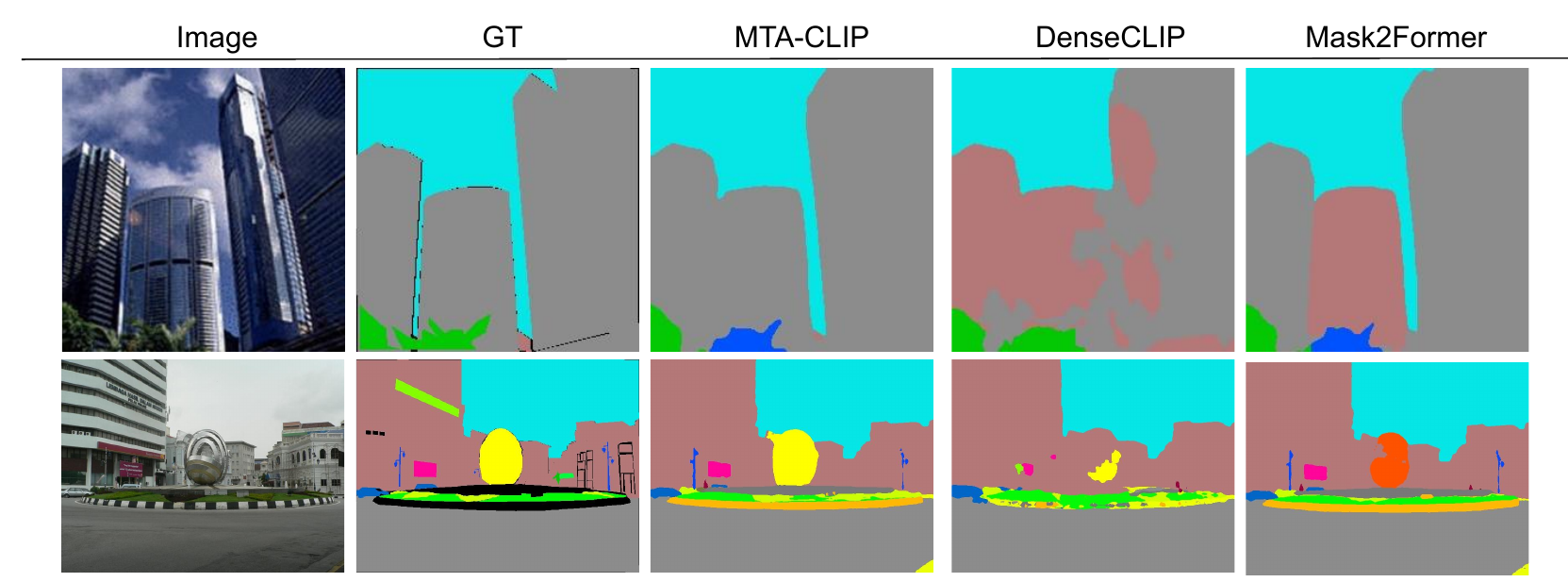}
\caption{Qualitative comparison. We compare our framework, MTA-CLIP's performance with Pixel-Text Alignment segmentation approach, DenseCLIP~\cite{rao2022denseclip} and Mask2Former~\cite{cheng2022masked} baseline with ResNet-50 backbone.}
\label{fig:qualitative}
\end{figure}

\subsection{Qualitative Analysis}
We present the qualitative comparison of our framework in~\cref{fig:qualitative}. We compare our framework with Pixel-to-Text Alignment method DenseCLIP~\cite{rao2022denseclip} and mask-based semantic segmentation baseline Mask2Former~\cite{cheng2022masked} for ResNet-50 backbone on ADE20k dataset. Overall, our framework segments the classes better than DenseCLIP and Mask2Former. Additional results with ResNet-101 backbone and on Cityscapes dataset is in supplement (Sec. 2).

\section{Conclusion}
This work introduces a novel framework, MTA-CLIP, designed for the semantic segmentation task, which implements innovative mask-level vision-language alignment. The proposed framework enhances mask representations by incorporating and aligning them with class-specific language features. Additionally, multiple context-specific prompts are employed to capture diverse representations of the mask segments. The framework is systematically evaluated using various backbones and pretraining strategies, consistently surpassing prior state-of-the-art approaches across diverse dataset benchmarks in the semantic segmentation task. We hope that this work stimulates further innovation in this challenging yet promising research direction.
\bibliographystyle{splncs04}
\bibliography{main}
\end{document}